\documentclass[runningheads]{llncs}
\usepackage[style=numeric, sorting=none]{biblatex}
\usepackage{graphicx}
\usepackage{colortbl}
\usepackage{nicematrix}
\usepackage{subcaption}
\usepackage{float}
\usepackage{caption}
\usepackage{lipsum}
\usepackage{placeins}

\begin{document}
\title{Rail Crack Propagation Forecasting Using Multi-horizons RNNs}
%
%
\author{Sara Yasmine OUERK \inst{1}
\and Olivier VO VAN \inst{2}
\and Mouadh YAGOUBI \inst{1} 
}
\authorrunning{S.Y. OUERK et al.}
%
\institute{IRT SystemX \and SNCF}

\maketitle              
\begin{abstract}
The prediction of rail crack length propagation plays a crucial role in the maintenance and safety assessment of materials and structures. Traditional methods rely on physical models and empirical equations such as Paris’ law, which often have limitations in capturing the complex nature of crack growth. In recent years, machine learning techniques, particularly Recurrent Neural Networks (RNNs), have emerged as promising methods for time series forecasting. They allow to model time series data, and to incorporate exogenous variables into the model.  
The proposed approach involves collecting real data on the French rail network that includes historical crack length measurements, along with relevant exogenous factors that may influence crack growth. First, a pre-processing phase was performed to prepare a consistent data set for learning. Then, a suitable Bayesian multi-horizons recurrent architecture was designed  to model the crack propagation phenomenon. 
Obtained results show that the Multi-horizons model outperforms state-of-the-art models such as LSTM and GRU.

\end{abstract}
\section{Introduction}

The French rail network has over 100,000 km of rail, including around 10,000 km for high-speed lines (LGV). The passage of rolling stock over these rails generates stresses in the rail, on the wheel-rail contact zone, which eventually leads to rolling contact fatigue. Defects resulting from this fatigue are monitored, and crack propagation is periodically checked, as a defect can propagate over several decades or a few months. When the length or depth of the crack becomes critical, it is imperative to correct the defect, otherwise there is a risk of rail break and potential derailment. Rolling contact fatigue is thus separated into two distinct phases, first the crack initiation, and then the crack propagation. In this paper, we focus on the latter and propose to build a predictive model that allows  to evaluate the residual life of an already existing crack before reaching the critical threshold. 
This phenomenon can be partially explained by physical models and many studies have been led to understand the impact of various parameters. Bonniot et al. showed that the crack propagation in the rail is complex and follow mixed non proportional propagation modes \cite{bonniot2018mixed}. Crack propagation speed depends on Stress Intensity Factor (SIF) identified from laboratory experiment, plastic deformation, friction between crack lips, its wear and corrosion and many other geometrical parameters such as initial crack width and direction, as shown by Fang et al. \cite{fang2022influence}. Moreover, other parameters in-situ are known to have an impact, such as track flexibility or acceleration and breaking and others still not quantified such as material decay over time. 
To deal with the lack of representativity of physical simulation in crack propagation modeling, we need to consider other  parameters and phenomena that can lead to a more and more computationally expensive simulations, prohibiting thus their use to solve real world problems.   
At the same time, the mass of real data   collected on various characteristics such as "infrastructure" and "traffic" makes it possible to investigate the potential of data-driven models. The problem can be seen as a time series forecasting of the crack length. In this paper, we propose a multi-horizon approach to predict the propagation of rail crack based on historical data that we compare with state of the art time series machine learning methods. The remainder of this paper is organized as follows. In section \ref{sec:rel_work} we present some recent related works. Section \ref{sec:data_process} describes the data processing analysis required to build the different models that are presented in section  \ref{sec:method}. The comparative results are discussed in Section \ref{sec:results} , and
as usual Section \ref{sec:conclusion}  summarizes the contribution of this work and suggests directions for future research.

\section{Related work}
\label{sec:rel_work}
Time series forecasting is a fundamental task in various domains, encompassing finance, weather prediction, demand forecasting, and more. Over the years, traditional and deep learning models have played a pivotal role in advancing the accuracy and effectiveness of time series forecasting.\\
Traditional approaches for time series forecasting have been widely used especially for univariate time series forecasting.
Holt et al. introduced a method commonly employed for time series forecasting, Exponential Smoothing (ES) \cite{holt1957forecasting}. They involve recursively updating the forecasted values by assigning exponentially decreasing weights to past observations. Simple Exponential Smoothing \cite{gardner1985exponential}, Holt's Linear Exponential Smoothing \cite{kalekar2004time}, and Holt-Winters' Seasonal Exponential Smoothing \cite{gardner1989note} are variations of this approach.\\
Autoregressive Integrated Moving Average (ARIMA) \cite{bartholomew1971time} is also a popular method for time series forecasting. It models the time series as a combination of autoregressive (AR), differencing (I), and moving average (MA) components. ARIMA models are widely used for stationary time series data.\\
These traditional approaches have been widely used in time series forecasting and have provided valuable insights in various domains. However, they have certain limitations that can impact their effectiveness and accuracy. In fact, many traditional time series forecasting methods assume that the underlying data follows a stationary process, where the statistical properties remain constant over time. However, real-world data often exhibits non-stationarity, such as trends, seasonality, and changing statistical properties. Failing to account for non-stationarity can lead to inaccurate forecasts. Moreover, these methods primarily focus on historical time series data and may not naturally incorporate external factors. However, many forecasting problems benefit from including additional variables, such as weather data. \\
While traditional time series forecasting approaches have their limitations, recent advancements in machine learning, such as deep learning models aim to address some of these challenges and provide more accurate and flexible forecasting capabilities. \\
Neural Networks (NN) have been widely used for time series forecasting and have achieved state-of-the-art performance in many applications. Neural networks, especially recurrent neural networks (RNNs) and their variants, have proven to be effective in capturing temporal dependencies and patterns in time series data. Moreover, there have been efforts to incorporate external factors or exogenous variables into time series forecasting models. These factors can include contextual information or additional time series that may influence the target variable.\\
One of the most popular RNN architectures for time series forecasting is the Long Short-Term Memory (LSTM) network \cite{hochreiter1997long}. LSTMs are designed to address the vanishing gradient problem and are capable of learning long-term dependencies in sequential data. They have been successfully applied to various time series forecasting tasks, including stock market prediction, energy load forecasting, and weather forecasting. \\
In recent years, other advanced variants of RNNs, such as Gated Recurrent Units (GRUs) \cite{cho2014learning} and Transformers \cite{vaswani2017attention}, have also shown promising results in time series forecasting. GRUs are similar to LSTMs but have a simpler architecture, which makes them computationally more efficient. \\
Transformers, originally introduced for natural language processing tasks, have been adapted for time series forecasting by leveraging self-attention mechanisms. Transformers have the advantage of parallel processing and have shown competitive performance in several domains.

\section{Data description and processing}
\label{sec:data_process}

Collected real data can be divided in four different categories. Each time it was possible, categorical data were converted to numerical data.
\begin{itemize}
    \item{\textbf{Infrastructure data}}: These data correspond to the network description. The interesting features to consider are all parameters that can change the vehicle dynamic, namely the rail linear mass, to take into account rail profile and vertical flexibility, sleeper type, rail grade, radius of curvature, cant, slope and side of the rail (left or right);
    \item{\textbf{Traffic data}}: These data correspond to the use of network. The dynamic impact of rolling stock is considered by maximal velocity allowed and quantity and number of acceleration and breaking. The rail loading is considered using annual tonnage (number of ton of vehicle seen by the rail) and number and type of vehicle (passenger or goods);
    \item{\textbf{Environment data}}: These are data not related to railway environment. The only environment data used here are temperatures and rain classified by type (low rain, strong storm, ice, snow, ...)
    \item{\textbf{Defect}}: These data correspond to the state of the network. Here, three different defects were selected, which represent most of rail defects in french railway, namely squats (in three different parts of the rail). Each defect is discovered at a recorded date and regularly visited to check its evolution. Each time, parameters such as crack length and measurement date are recorded. 
\end{itemize}
One last parameter is considered and called "UIC Group". It is strongly correlated with speed limit and tonnage and defines maintenance conditions. Through this parameter are thus included other unavailable data at the time of the study such as grinding works. 
These data present a number of anomalies (inconsistent format, missing values, etc.), which necessitated a data preprocessing phase to obtain a consistent database  to train the Machine Learning models.
\\ \\
Note that crack data was the most challenging to process for several reasons:
\begin{itemize}
    \item Crack length values also present anomalies linked to database filling errors (negative values, exceeding certain thresholds, or considerable falls in values);
    \item Discovery date happened between 2008 and 2018 and crack life before it is removed can vary from several months to several years;
    \item Visit dates at which the crack length is measured are manually and empirically planed, the duration between two visits can thus vary from one week to a couple of years;
    \item The perceived high risk cracks are frequently visited and lead to sequence length (the time series) longer than others;
    \item Abrupt propagation have been observed for some defects. This behavior may be physically explained (caused by an extremely cold day) , or simply based on a human judgement to merge two spatially close defects;
    \item Abrupt reduction of the crack length, which can be due to rail grinding;
    \item Measurement uncertainty, which is a known issue and led to approximate the measured length to the closest multiple of 5.
\end{itemize} 
\textbf{Data  processing} \\ 
All the above information have been crossed to create a single training dataset containing all the information. 
The anomalies mentioned above were also addressed based on experts knowledge on the data.
To overcome the problem of irregular time steps in the time series, an interpolation was performed. A frequency of 3 months was chosen and a linear average was computed on all series, resulting in 3-month time-step series with a maximum length of 59 time steps. After this step, defects with a fall in values greater than $15mm$ were removed from the database, to avoid introducing errors into the learning model. Drops in values of less than $15mm$ are tolerated, as it is possible to have variations in measurement conditions such as temperature variation that can lead to crack closure and reduce the size measured as explained in \cite{kou2021fully}. The measurement is also subject to operator interpretation of the observed signal and can thus vary with operators.  \\ \\
\textbf{Feature extraction} \\ 
In the collected data, defect discovery dates vary widely, with some defects being more recent than others. To consider this information in the learning process, we set up an input variable that calculates the elapsed time since the defect was discovered. \\
The crack propagation speed was also calculated between time steps, which can give an indication of how fast the crack length propagates in a given context for the learning model. This information can only be used in the past horizon (the notion of horizon will be introduced in section \ref{sec:horizon}) and not in the future horizon, to avoid giving information on the lengths to be predicted. This feature extraction and selection resulted in 37 exogenous features for each time step in the time series.

\section{Modeling approaches} 
\label{sec:method}
 
\subsection{Feature based modeling } %
Initially, crack length values are considered unknown to the model. Only exogenous variables will be taken into account by the model to predict the corresponding crack lengths.
As mentioned in the previous section, several variables are available. The time series are therefore multivariate, with several dynamic (evolving over time) or static features for the different time steps.
For this configuration, sequences were created using a sliding window of size $t$. \\ \\
The goal is to model the distribution of the crack length sequence, knowing its current context $X_{1:t}$, as 
\begin{equation}
    P(Y_{1:t}|X_{1:t}).
\end{equation}
Were $X_{1:t}$ represents the exogenous feature (static and dynamic) by time step, and $Y_{1:t}$ their corresponding crack length values to be predicted.\\
Static features are encoded using Fully Connected (FC) layers,  the dynamical features are encoded also using Fully Connected (FC) layers and then passed to one type of recurrent layers (RNN, LSTM or GRU) which can handle the time dependency between time steps. These models are respectively called \textbf{RNN-FC}, \textbf{LSTM-FC} and \textbf{GRU-FC}.
\subsection{Considering the historical crack length values}
\label{sec:horizon}
\begin{figure}[h!]
    \centering
    \includegraphics[scale=0.8]{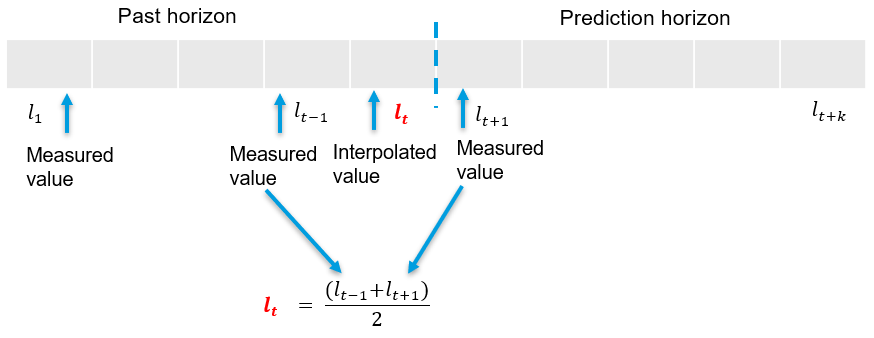}
     \caption{Example of interpolation for the last step before the prediction horizon}
    \label{fig1}
\end{figure}

For this new setup, a dataset containing crack length sequences was created using a sliding window of length $t + k$ over our time series. Each position of the sliding window contains a sample in our dataset with the first $t$ values of crack lengths (the history), and their corresponding contextual features being the input of the past horizon and the last $k$ values (the forecasting horizon) being the output. \\
The goal in this case is to model the distribution of the crack length sequence, knowing its historical features $X_{1:t}$ and measurements $Y_{1:t}$, as
\begin{equation}
    P(Y_{t+1:t+k}|Y_{1:t}, X_{1:t}).
\end{equation}
As mentioned above, interpolation is used to deal with the problem of irregular time series. The interpolated values are calculated using a linear average. For some time series, the past horizon may contain interpolated length values after the last measured value. These values are calculated using crack length values from the prediction horizon, as explained in figure \ref{fig1}. So, introducing them to the learning model will give information about the future values that are supposed to be unknown for the model, and thus may introduce a bias for the learning process.\\
To avoid this problem, only interpolated values before the last measured value are included. For time steps interpolated after this step, the last measured value is used to replace the interpolated steps.
As an example in table \ref{tab1}, we assume that crack length values of the defect corresponding to the past horizon are the values in the first row. The "Last measured value" variable indicates the last measured crack value (not an interpolated value), the "Step is interpolated" variable indicates whether the time step corresponds to an interpolated or non-interpolated (measured) crack length value.\\
The fifth time step is interpolated, and is the last time step before the prediction horizon, so its value can give information about the first value in the prediction horizon. Consequently, this value is replaced by the last measured crack length value. The model input for the "historical crack length values" feature will then be the "model input" variable in the table. \\ 
It should be noted that the model will be less accurate with this modification, but at least it will avoid biasing it with information it is not supposed to know. Some variables have been added to indicate whether the time step is interpolated and, if so, the number of time steps since the last measurement. This will reduce the effect of this replacement on performance.
\begin{table}[h]
   \caption{Example of model input of crack length values in the past horizon with last crack length value replacement}
 \centering
\begin{tabular}{ |>{\columncolor{gray!15}}c|c|c|c|c|>{\columncolor{blue!15}}c| } 
\hline
Crack length & 30 & 32.5 & 35 & 35 & \textbf{38.125} \\
\hline
Last measured value & 30 & 30 & 35 & 35 & \textbf{35} \\ 
 \hline
Time step is interpolated & No & Yes & No & No & \textbf{Yes}  \\ 
\hline
Model input & 30 & 32.5 & 35 & 35 & \textbf{35}   \\ 
\hline
\end{tabular} \newline
\label{tab1}
\end{table}
\FloatBarrier 
\subsubsection{Simple Recurrent model}\hfill \\
For this model, only historical exogenous characteristics and corresponding crack length values are considered. These variables are passed on to the recurrent layer (LSTM/GRU), then their latent representation is passed on to some fully connected layers in order to infer crack length values in the future. These models are called \textbf{LSTM-FC-LH and GRU-FC-LH}, where LH refers to the historical crack lengths.
\subsubsection{Multi-horizons recurrent model }
\begin{figure}[h!]
    \centering
    \includegraphics[scale=0.8]{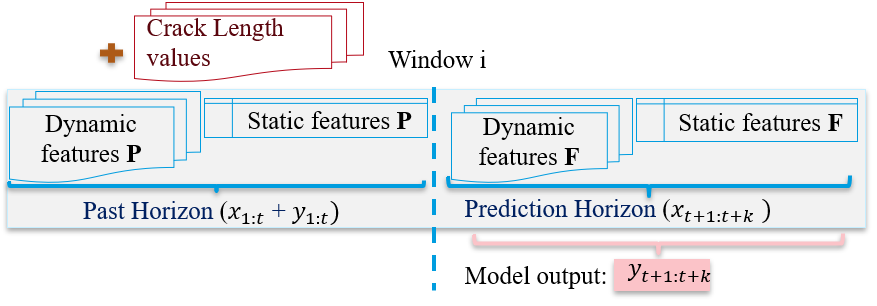}
    \caption{Scheme of the prediction model}
    \label{fig2}
\end{figure}

\begin{figure}[h]
    \centering
    \includegraphics[scale=0.5]{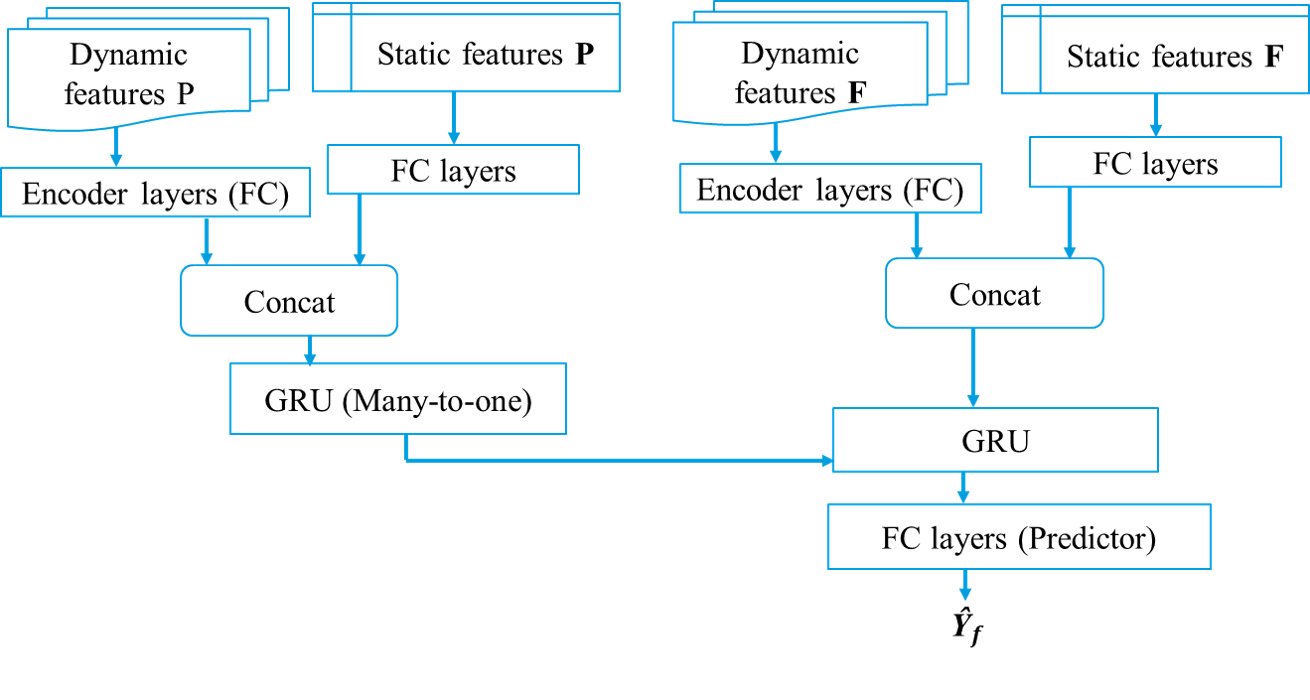}
    \caption{Architecture of the multi-horizons recurrent model}
    \label{fig3}
\end{figure}

\hfill \\
In a second step, a model was implemented to consider both historical context $ X_{1:t}$ and lengths $Y_{1:t}$, as well as the current context $X_{t+1:t+k}$. The aim is to model the distribution,
\begin{equation}
    P(Y_{t+1:t+k}|Y_{1:t}, X_{1:t},X_{t+1:t+k}).
\end{equation} 
\noindent This model is a recurrent neural network with multiple time horizons. It consists of a past horizon which takes as input exogenous variables and historical crack length measurements, and a future prediction horizon which takes as input the encoded output from the past horizon as well as current contextual variables in order to infer future crack length values, as described in Figure \ref{fig2}. \\ 
The general architecture of the multi-horizon model is shown in Figure \ref{fig3}.\\
\noindent For all the described models above, a customized Mean Squared Errors ($MSE$) has been used for learning. This loss is an $MSE$ loss that ignores the padded time steps in order to avoid introducing bias to the model. 
\subsubsection{Bayesian Multi-horizons recurrent model }\hfill \\ 
As mentioned above, crack length measurements are subject to uncertainty. This uncertainty is related to the data quality that cannot be reduced by adding more data, but it can be quantified. This type of uncertainty is called $Aleatoric$ uncertainty and captures inherent noise in the observations. The learning model itself may be also uncertain regarding its predictions, due to a lack of learning data for example. This is called $epistemic$ uncertainty and can be reduced  by observing more data.\\
\noindent The multi-horizons model described above has been adapted, based on a Bayesian approach suggested by Kendall et al. \cite{kendall2017uncertainties}, to allow uncertainty estimation in parallel with model prediction. This model is called the Bayesian Multi-horizons model (B-MH).\\
The B-MH model output, is composed of  predictive mean $\hat{y}$ as well as predictive variance $\hat{\sigma}^2$.\\
The general architecture of the model remains unchanged, with only the last fully connected layers duplicated in order to output both $\hat{y}$ and $\hat{\sigma}^2$. $\hat{y}$ represents the predictive mean crack length and $\hat{\sigma}^2$ its predictive variance.\\
A Gaussian likelihood is used to model the aleatoric uncertainty, as the available crack length values  follow a Gaussian distribution. This induces the minimization loss function for a given sequence $x_i$,\\
\begin{equation}
  L_\mathrm{B\_MH}\ =\ \frac{1}{N_i} \sum_{j=1}^{N_i} \frac{1}{2\hat{\sigma}(x_{ij})^2} ||y_{ij}-\hat{y}_{ij}||^2 +  \frac{1}{2} log(\hat{\sigma}(x_{ij})^2),
\end{equation}
were $\hat{\sigma}(x_{ij})^2$ is the predictive variance for the the time step $j$ of the sequence $x_i$, $\hat{y}_{ij}$ its predictive mean and $N_i$ the number of time steps in the sequence $x_i$.\\
The variance $\hat{\sigma}^2$ is implicitly learnt from the loss function. The division of the residual loss $||y_{ij}-\hat{y}_{ij}||^2$ (which represent the $MSE$ loss) by $\hat{\sigma}(x_{ij})^2$ makes the model more robust to noisy data. In fact, data for which the model has learned to predict a high uncertainty will have lower effect on loss. The second regularization term prevents the network from predicting infinite uncertainty.\\
For numerical stability, and to avoid dividing by zero or either predicting negative variance, the term $\hat{\sigma}(x_{ij})^2$ is replaced by the term $s_{ij} = log (\hat{\sigma}(x_{ij})^2)$. The weights of the two terms in the equation have been set to $\frac{2}{3}$ and$\frac{1}{3}$ respectively, to give more weight to the $MSE$ than to the regularization term, resulting in the minimization function,
\begin{equation}
 L_\mathrm{B\_MH}\ =\ \frac{1}{N_i} \sum_{j=1}^{N_i} \frac{2}{3} exp(- s_{ij}) ||y_{ij}-\hat{y}_{ij}||^2 +  \frac{1}{3} s_{ij}.
\end{equation}
To quantify the uncertainty, a dropout approach \cite{srivastava2014dropout} is used as Bayesian approximation. The model is trained with dropout before every weight layer. Contrary to what is usually done for a network trained with dropout layers, dropout remains activated during inference to generate stochastic rather than deterministic outputs. $T$ stochastic prediction samples are performed using Dropout, allowing to approximate the predictive uncertainty for one observation as
\begin{equation}
\mathrm{Var}(y)\ \approx\ (\frac{1}{T}\sum_{t=1}^{T} \hat{y}_{t} - (\frac{1} {T}\sum_{t=1}^{T} \hat{y}_{t})^2 ) + \frac{1}{T} \sum_{t=1}^{T} \hat{\sigma}_{t}^2,
\label{eq6}
\end{equation}
with $\{{ \hat{y}_{t}, \hat{\sigma}_{t}^2 }\}_{t=1}^{T}$ the set of $T$ sampled outputs after each forward pass.\\
The first term of this total variance corresponds to the epistemic uncertainty and the second one corresponds to the aleatoric uncertainty.	

\section{Experiments}
\label{sec:results}
\subsection{Data preparation for Learning}
Whatever the model used for learning, the generated time series have been pre-processed to ensure that the learning models function correctly.\\
By choosing a maximum size for the prediction horizon at a given value, not all the series generated have the same length. As some are shorter than the maximum length, these series have been completed by adding zeros at the end, so that they all have the same length. These completed time steps will be ignored when calculating the cost functions by implementing custom functions that ignore these time steps for backpropagation.\\
After this step, the dataset is divided into three parts: 60 \%  training set for the learning procedure, 20 \% validation set for hyperparameter optimization and convergence control, and 20 \% test set for performance evaluation.
The division strategy adopted ensures that the subsequences of a given defect series belong to only one of the three previous sets.\\
The time series are then normalized using a custom time series standard scaler, so that their mean is 0 and their standard deviation is 1. This makes the model much more robust to outliers. Min-max normalization has also been tested, but gives slightly poorer results.

\subsection{Settings}
The work has been implemented in Python using Pytorch. All the experiments are conducted using an  NVIDIA A40 GPU.\\
Adam optimizer is used to perform the gradient descent minimization of the loss function. The activation function used is the $Tanh$ function for all hidden layers.\\
The convergence of the models is checked on learning rates from $10^{-1}$ to $10^{-4}$,  and on different batch sizes. The models perform best with the learning rate of 0.001 and batch size of $128$. The models are also fitted over a variable number of epochs, the classical recurrent models converge after about $25$ epochs, and the multi-horizons models converge after  $10$ epochs. \\
To benchmark the different models, many ML and physical metrics are used to compare their performances. $MAE$ and $RMSE$ errors are used as machine learning metrics.
Other physical criteria are considered to avoid some physical constraints violations such as the drop in the crack length, a phenomenon that should not occur physically (the crack can either progress or remain constant). These physical criteria are:
\begin{itemize}
\item{\textbf{MSQNS}}, for Mean SeQuence Negative Slope, is the percentage of sequences that contains at least one fall in the predicted values;
\item{\textbf{MSTNS}}, for Mean STeps Negative Slope, is the percentage of steps that contains at least one fall in the predicted values. 
\item{\textbf{MLNS}}, for Mean Length Negative Slope, is the mean value of the fall in predicted length values. As a reminder, the observation time series themselves contain drops in values of up to 15mm. 
\end{itemize}
The computation of evaluation criteria for all reported experiments in this paper is performed using the recently proposed LIPS Framework for benchmarking learned physical systems \cite{leyli2022lips}.
\subsection{Experiments with simple configuration (without historical crack length values)}
For this modeling, there is no notion of horizons in the generation of sequences. Generated sequences are of size 4 (we need to anticipate crack lengths values over a period of one year with a time step of 3 months). As previously stated, only exogenous variables are considered for prediction.
Recurrent models were compared using the various ML and physical criteria defined above. This comparison is made in particular for the average score over the 4 time steps to be predicted (mean MAE and mean RMSE), as well as for the scores linked to the prediction of the first time step (MAE $1^{st}$ and RMSE $1^{st}$) as shown in Table \ref{tab2}. The results show that the GRU-FC model outperforms LSTM-FC and RNN-FC in terms of machine learning criteria. The LSTM-FC and RNN-FC models have quite similar ML results, but the LSTM-FC model gives the best results in terms of physical criteria.
\begin{table}[h!]
\centering
\caption{ML and Physical results for the recurrent models without using historical crack length values}
\begin{tabular}{c|
>{\columncolor[HTML]{FFFFFF}}c |
>{\columncolor[HTML]{FFFFFF}}c |
>{\columncolor[HTML]{FFFFFF}}c |
>{\columncolor[HTML]{FFFFFF}}c |
>{\columncolor[HTML]{FFFFFF}}c |
>{\columncolor[HTML]{FFFFFF}}c |
>{\columncolor[HTML]{FFFFFF}}c }
\hline
Model                                                 & \cellcolor[HTML]{EFEFEF}{\color[HTML]{343434} \begin{tabular}[c]{@{}c@{}}MAE \\ $1^{st}$\end{tabular}} & \cellcolor[HTML]{EFEFEF}{\color[HTML]{343434} \begin{tabular}[c]{@{}c@{}}Mean \\ MAE\end{tabular}} & \cellcolor[HTML]{EFEFEF}{\color[HTML]{343434} \begin{tabular}[c]{@{}c@{}}RMSE \\ $1^{st}$\end{tabular}} & \cellcolor[HTML]{EFEFEF}\begin{tabular}[c]{@{}c@{}}Mean \\ RMSE\end{tabular} & \cellcolor[HTML]{EFEFEF}{\color[HTML]{343434} MLNS} & \cellcolor[HTML]{EFEFEF}{\color[HTML]{343434} MSQNS} & \cellcolor[HTML]{EFEFEF}MSTNS       \\ \hline
\cellcolor[HTML]{C0C0C0}{\color[HTML]{000000} RNN-FC}  & {\color[HTML]{000000} 10.48}                                                                      & {\color[HTML]{000000} 10.47}                                                                       & {\color[HTML]{000000} 13.67}                                                                       & {\color[HTML]{000000} 13.66}                                                 & {\color[HTML]{000000} 1.72}                         & {\color[HTML]{000000} 29\%}                          & {\color[HTML]{000000} 8\%}          \\ \hline
\cellcolor[HTML]{C0C0C0}{\color[HTML]{000000} GRU-FC}  & {\color[HTML]{000000} \textbf{9.65}}                                                              & {\color[HTML]{000000} \textbf{9.45}}                                                               & {\color[HTML]{000000} \textbf{12.60}}                                                              & {\color[HTML]{000000} \textbf{12.38}}                                        & {\color[HTML]{000000} 2.59}                         & {\color[HTML]{000000} 24\%}                          & {\color[HTML]{000000} 6\%}          \\ \hline
\cellcolor[HTML]{C0C0C0}{\color[HTML]{000000} LSTM-FC} & {\color[HTML]{000000} 10.54}                                                                      & {\color[HTML]{000000} 10.53}                                                                       & {\color[HTML]{000000} 13.75}                                                                       & {\color[HTML]{000000} 13.72}                                                 & {\color[HTML]{000000} \textbf{1.18}}                & {\color[HTML]{000000} \textbf{3\%}}                  & {\color[HTML]{000000} \textbf{1\%}} \\ \hline
\end{tabular}
\label{tab2}
\end{table}
\FloatBarrier 
\subsection{Experiments considering historical crack length values}
\subsubsection{Experiments with recurrent models} \hfill\\
For this modeling, time series were created using a sliding window of size 9: with a past horizon of size 5 and a prediction horizon of size 4. The size of the past horizon containing historical crack values was chosen at 5 time steps, inspired by \cite{lara2021experimental} which suggests that a past horizon of size $1.25 \times k$ ($k$ being the size of the prediction horizon) gives the best prediction results. \\
Table \ref{tab3} shows ML and physical criteria for the recurrent models that considers historical crack length values. ML scores include the MAE for the different time steps in the prediction horizon (from $t+1$ to $t+4$) and their average value, and the RMSE score for the first time step in the the prediction horizon and the average score over the entire prediction horizon. The LSTM-FC-LH model gives slightly better results than the GRU-FC-LH. For the physical criteria, this time it is the GRU-FC-LH model that gives slightly better results.
\begin{table}[h!]
\centering
\caption{ML and Physical results for the recurrent models considering historical crack length values}
\begin{tabular}{c|c|c|c|c|c|c|c|c|c|c}
\hline
\rowcolor[HTML]{EFEFEF} 
\cellcolor[HTML]{FFFFFF}Model                             & {\color[HTML]{343434} \begin{tabular}[c]{@{}c@{}}MAE \\ 1\end{tabular}} & \begin{tabular}[c]{@{}c@{}}MAE \\ 2\end{tabular} & \begin{tabular}[c]{@{}c@{}}MAE \\ 3\end{tabular} & \begin{tabular}[c]{@{}c@{}}MAE  \\ 4\end{tabular} & {\color[HTML]{343434} \begin{tabular}[c]{@{}c@{}}Mean \\ MAE\end{tabular}} & {\color[HTML]{343434} \begin{tabular}[c]{@{}c@{}}RMSE \\ $1^{st}$\end{tabular}} & \begin{tabular}[c]{@{}c@{}}Mean \\ RMSE\end{tabular} & {\color[HTML]{343434} MLNS}          & {\color[HTML]{343434} MSQNS}          & MSTNS                                  \\ \hline
\rowcolor[HTML]{FFFFFF} 
\cellcolor[HTML]{C0C0C0}{\color[HTML]{000000} LSTM-FC-LH} & {\color[HTML]{000000} \textbf{2.37}}                                    & \cellcolor[HTML]{FFFFFF}\textbf{3.05}            & \cellcolor[HTML]{FFFFFF}\textbf{3.85}            & \cellcolor[HTML]{FFFFFF}\textbf{4.51}             & {\color[HTML]{000000} \textbf{3.45}}                                       & {\color[HTML]{000000} \textbf{4.72}}                                       & {\color[HTML]{000000} \textbf{6.01}}                 & {\color[HTML]{000000} 1.16 mm}       & {\color[HTML]{000000} 1\%}            & {\color[HTML]{000000} 0.15\%}          \\ \hline
\rowcolor[HTML]{FFFFFF} 
\cellcolor[HTML]{C0C0C0}{\color[HTML]{000000} GRU-FC-LH}  & {\color[HTML]{000000} \textbf{2.37}}                                    & \cellcolor[HTML]{FFFFFF}3.11                     & \cellcolor[HTML]{FFFFFF}\textbf{3.85}            & \cellcolor[HTML]{FFFFFF}4.58                      & {\color[HTML]{000000} 3.49}                                                & {\color[HTML]{000000} 4.77}                                                & {\color[HTML]{000000} 6.06}                          & {\color[HTML]{000000} \textbf{1.07}} & {\color[HTML]{000000} \textbf{0.5\%}} & {\color[HTML]{000000} \textbf{0.13\%}} \\ \hline
\end{tabular}
\label{tab3}
\end{table}
\subsubsection{Experiments with the Multi-horizons and Bayesian Multi-horizons models} 
\begin{figure}[h!]
    \centering
    \includegraphics[scale=0.45]{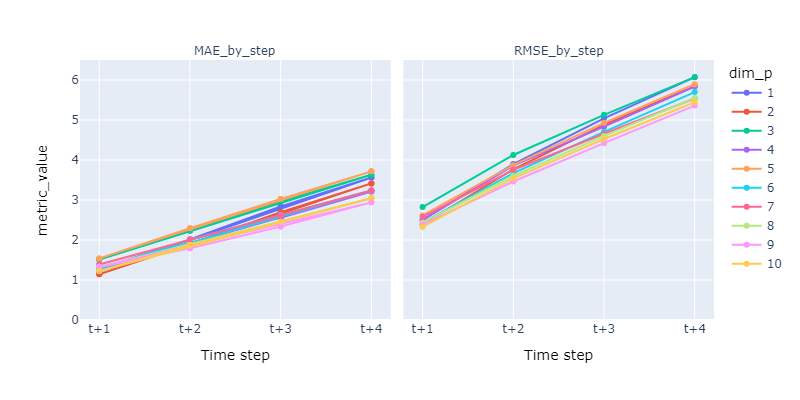}
    \caption{MAE and RMSE scores for the prediction horizon using the multi-horizons model with different past horizon lengths.}
    \label{fig4}
\end{figure}

\begin{figure}[h!]
    \centering
    \includegraphics[scale=0.45]{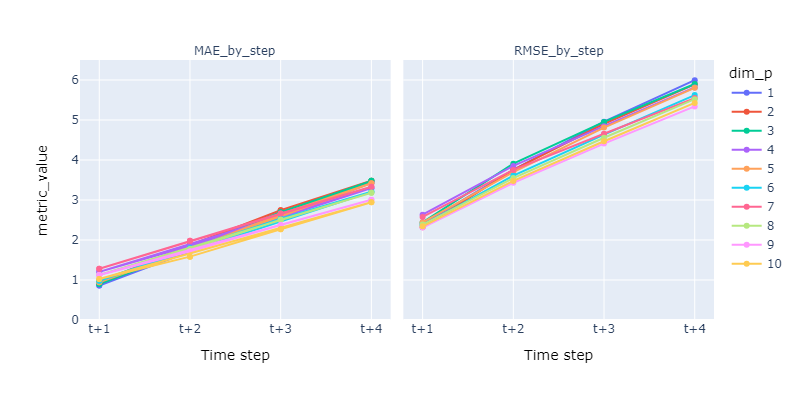}
    \caption{MAE and RMSE scores for the prediction horizon using the Bayesian multi-horizons model with different past horizon lengths}
    \label{fig5}
\end{figure}

\hfill\\
For this modelling, a number of past horizon sizes were tested to see their effect on the various criteria to be minimized.\\
\noindent Tables \ref{tab4} and \ref{tab5} show the results of the different ML and physical criteria of the multi-horizons model and the Bayesian multi-horizons model with different past horizon sizes. Good results can already be obtained from a single measurement in the past horizon. The size of the training set decreases as the size of the past horizon increases, due to the filtering of sequences to respect the minimum size. The choice of the size of the past horizon is conditioned both by the criteria to be minimized as far as possible and by industrial use. Indeed, information on historical measurements is sometimes available for just 1 or 2 time steps, which corresponds to three months or less, but we still want to predict crack lengths in the future because some cracks might have exceeded the security threshold before 6 months. The model must therefore be able to make predictions even with a limited past horizon size. 
\begin{table}[h!]
\centering
\caption{ML and physical criteria results for the multi-horizons model considering different past horizons lengths for prediction}
\begin{tabular}{c|c|c|c|c|c|c|c|c}
\hline
\rowcolor[HTML]{EFEFEF} 
\cellcolor[HTML]{C0C0C0}dim\_hp & \begin{tabular}[c]{@{}c@{}}nb\_sequences\\      train\end{tabular} & \begin{tabular}[c]{@{}c@{}}MAE \\ $1^{st}$\end{tabular} & \begin{tabular}[c]{@{}c@{}}Mean \\ MAE\end{tabular} & \cellcolor[HTML]{EFEFEF}\begin{tabular}[c]{@{}c@{}}RMSE \\ $1^{st}$\end{tabular} & \begin{tabular}[c]{@{}c@{}}Mean \\ RMSE\end{tabular} & \begin{tabular}[c]{@{}c@{}}MSQNS\\      \%\end{tabular} & \begin{tabular}[c]{@{}c@{}}MSTNS\\     \%\end{tabular} & \begin{tabular}[c]{@{}c@{}}MLNS\\    mm\end{tabular} \\ \hline
\textbf{1}                      & 294018                                                             & 1.22                                               & 2.41                                                & 2.50                                                                        & 4.38                                                 & 2.95                                                    & 0.79                                                   & 1.14                                                 \\ \hline
\textbf{2}                      & 265519                                                             & 1.15                                               & 2.29                                                & 2.39                                                                        & 4.22                                                 & 2.85                                                    & 0.76                                                   & 1.13                                                 \\ \hline
\textbf{3}                      & 238222                                                             & 1.51                                               & 2.58                                                & 2.82                                                                        & 4.54                                                 & 4.58                                                    & 1.20                                                   & 1.18                                                 \\ \hline
\textbf{4}                      & 216021                                                             & 1.28                                               & 2.26                                                & 2.50                                                                        & 4.26                                                 & 14.09                                                   & 3.61                                                   & 1.11                                                 \\ \hline
\textbf{5}                      & 193901                                                             & 1.54                                               & 2.64                                                & 2.62                                                                        & 4.33                                                 & 2.87                                                    & 0.74                                                   & 1.13                                                 \\ \hline
\textbf{6}                      & 173598                                                             & 1.27                                               & 2.29                                                & 2.43                                                                        & 4.13                                                 & 6.58                                                    & 1.69                                                   & 1.08                                                 \\ \hline
\textbf{7}                      & 158040                                                             & 1.39                                               & 2.31                                                & 2.58                                                                        & 4.13                                                 & 4.80                                                    & 1.22                                                   & 1.17                                                 \\ \hline
\textbf{8}                      & 141407                                                             & 1.33                                               & 2.17                                                & 2.43                                                                        & 4.05                                                 & 8.61                                                    & 2.22                                                   & 1.08                                                 \\ \hline
\textbf{9}                      & 126847                                                             & 1.33                                               & 2.10                                                & 2.39                                                                        & 3.91                                                 & 6.78                                                    & 1.74                                                   & 1.16                                                 \\ \hline
\textbf{10}                     & 113175                                                             & 1.23                                               & 2.15                                                & 2.33                                                                        & 3.96                                                 & 14.91                                                   & 3.80                                                   & 1.14                                                 \\ \hline
\end{tabular}
\label{tab4}
\end{table}
\begin{table}[h!]
\centering
\caption{ML and physical criteria results for the Bayesian multi-horizons model (B-MH) considering different past horizons lengths for prediction}
\begin{tabular}{c|c|c|c|c|c|c|c|c}
\hline
\rowcolor[HTML]{EFEFEF} 
\cellcolor[HTML]{C0C0C0}dim\_hp & \begin{tabular}[c]{@{}c@{}}nb\_sequences\\      train\end{tabular} & \begin{tabular}[c]{@{}c@{}}MAE \\ $1^{st}$\end{tabular} & \begin{tabular}[c]{@{}c@{}}Mean \\ MAE\end{tabular} & \cellcolor[HTML]{EFEFEF}\begin{tabular}[c]{@{}c@{}}RMSE \\ $1^{st}$\end{tabular} & \begin{tabular}[c]{@{}c@{}}Mean \\ RMSE\end{tabular} & \begin{tabular}[c]{@{}c@{}}MSQNS\\      \%\end{tabular} & \begin{tabular}[c]{@{}c@{}}MSTNS\\     \%\end{tabular} & \begin{tabular}[c]{@{}c@{}}MLNS\\    mm\end{tabular} \\ \hline
\textbf{1}                      & 294018                                                             & 0.86                                               & 2.21                                                & 2.40                                                                        & 4.28                                                 & 1.99                                                    & 0.52                                                   & 1.09                                                 \\ \hline
\textbf{2}                      & 265519                                                             & 0.94                                               & 2.26                                                & 2.32                                                                        & 4.22                                                 & 1.51                                                    & 0.39                                                   & 1.14                                                 \\ \hline
\textbf{3}                      & 238222                                                             & 0.90                                               & 2.21                                                & 2.44                                                                        & 4.30                                                 & 1.05                                                    & 0.29                                                   & 1.13                                                 \\ \hline
\textbf{4}                      & 216021                                                             & 1.20                                               & 2.23                                                & 2.63                                                                        & 4.29                                                 & 3.56                                                    & 0.91                                                   & 1.05                                                 \\ \hline
\textbf{5}                      & 193901                                                             & 0.94                                               & 2.19                                                & 2.44                                                                        & 4.19                                                 & 1.30                                                    & 0.35                                                   & 1.09                                                 \\ \hline
\textbf{6}                      & 173598                                                             & 0.98                                               & 2.13                                                & 2.37                                                                        & 4.06                                                 & 2.87                                                    & 0.73                                                   & 1.04                                                 \\ \hline
\textbf{7}                      & 158040                                                             & 1.28                                               & 2.31                                                & 2.58                                                                        & 4.13                                                 & 1.41                                                    & 0.37                                                   & 1.07                                                 \\ \hline
\textbf{8}                      & 141407                                                             & 1.13                                               & 2.15                                                & 2.38                                                                        & 4.00                                                 & 9.83                                                    & 2.49                                                   & 1.04                                                 \\ \hline
\textbf{9}                      & 126847                                                             & 1.13                                               & 2.06                                                & 2.31                                                                        & 3.87                                                 & 6.20                                                    & 1.58                                                   & 1.02                                                 \\ \hline
\textbf{10}                     & 113175                                                             & 1.02                                               & 1.96                                                & 2.34                                                                        & 3.93                                                 & 2.94                                                    & 0.78                                                   & 1.09                                                 \\ \hline
\end{tabular}
\label{tab5}
\end{table}
\noindent Figure \ref{fig4} and \ref{fig5} show ML scores (MAE and RMSE) for both the multi-horizons and Bayesian multi-horizons models using different past horizons lengths, these scores are presented in detail over the entire prediction horizon. Models errors increase with distance from the past horizon. The Bayesian multi-horizons model outperforms the multi-horizons model over the entire forecast horizon.

\begin{figure}[h!]
  \centering
  \begin{subfigure}[b]{0.49\linewidth}
    \includegraphics[scale=0.5]{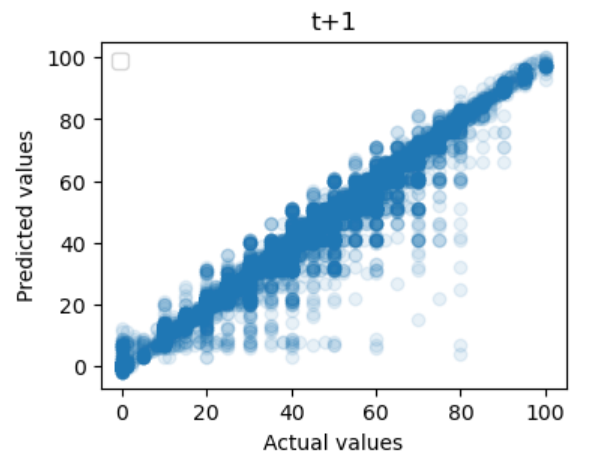}
    \caption{After 3 months}
  \end{subfigure}
  \begin{subfigure}[b]{0.49\linewidth}
    \includegraphics[scale=0.5]{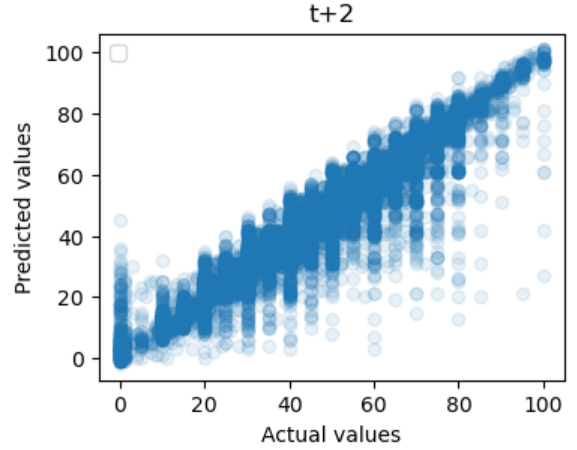}
    \caption{After 6 months}
  \end{subfigure}
  \begin{subfigure}[b]{0.49\linewidth}
    \includegraphics[scale=0.5]{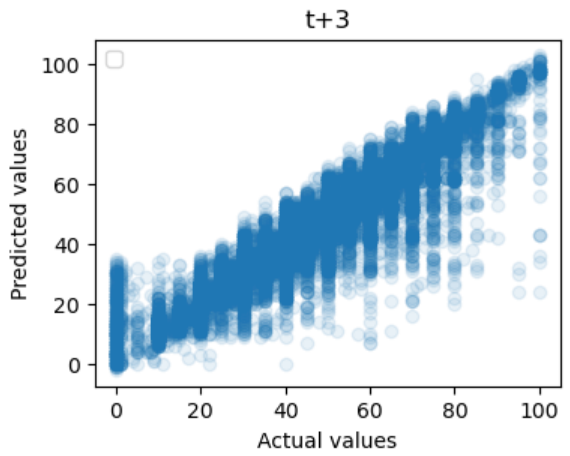}
    \caption{After 9 months}
  \end{subfigure}
  \begin{subfigure}[b]{0.49\linewidth}
    \includegraphics[scale=0.5]{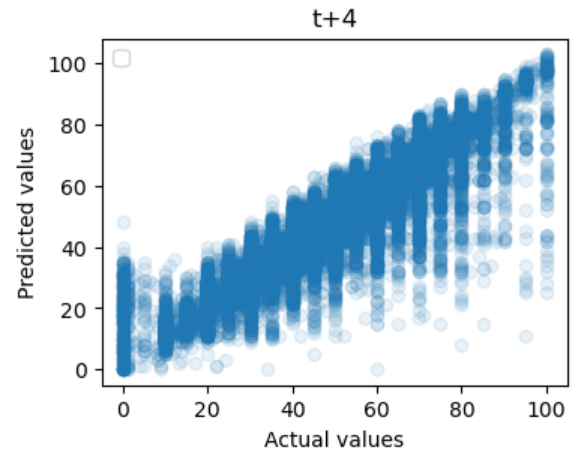}
    \caption{After 1 year }
  \end{subfigure}
  \caption{Actual vs predicted crack length values over the prediction horizon}
  \label{fig6}
\end{figure}

\noindent Figure \ref{fig6} shows the scatter plots for each time step in the prediction horizon. The x-axis and the y-axis correspond to the measured and predicted values of
crack length respectively. There is a high density around the $y=x$ line which explains the good prediction scores. There are, however, some miss-predicted values, especially when crack lengths become large, where the model tends to underestimate them. This result is mainly due to the small percentage of large crack length values in the dataset.

\subsubsection{Uncertainty quantification using the Bayesian multi-horizons model} 
\begin{figure}[]
  \centering
  \begin{subfigure}[b]{0.495\linewidth}
    \includegraphics[width=\linewidth]{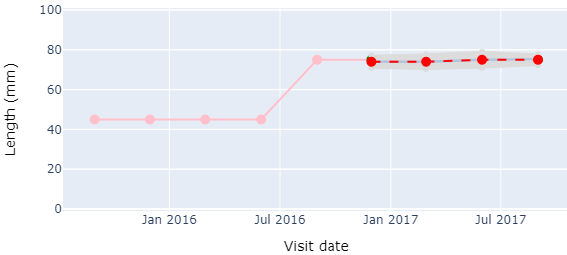}
    \caption{Example 1}
  \end{subfigure}
  \begin{subfigure}[b]{0.495\linewidth}
    \includegraphics[width=\linewidth]{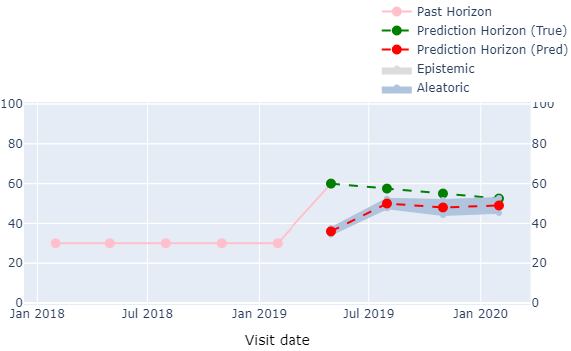}
    \caption{Example 2}
  \end{subfigure}
  \caption{Crack length propagation example with corresponding uncertainty estimation using the Bayesian Multi-horizons model}
  \label{fig7}
\end{figure}

As described above, uncertainty quantification is performed after the training of the model using Monte Carlo dropout sampling. Dropout is set after each layer (except the last one) and 50 Monte Carlo samples were generated for each time series. Then, the sum of the two types of uncertainty is calculated using equation \ref{eq6}. The dropout rate was varied from 10\% to 50\%. Aleatoric uncertainty does not vary widely, as it is linked to the inherent noise of the data. Epistemic uncertainty, on the other hand, increases as the dropout rate is increased, since it is linked to the learning model. As a result, total uncertainty increases, as does the size of the confidence interval, resulting in higher coverage.
For the rest of this study, a dropout rate of 10\% is set after each layer, and an approximate $95\%$-level prediction confidence interval is constructed. Results show that only $48\%$ of time steps are covered by this confidence interval. Indeed, as mentioned above, all the measured length values were approximated to the closest multiple of $5$, which led us to add  threshold of $5$ to the confidence interval. This time, about $93\%$ of time steps are covered by the new confidence interval.\\
Figure \ref{fig7} shows some example of crack length propagation, the corresponding predicted values and uncertainty estimated values. Example 1 is a case of a crack whose final value becomes significant (around $80mm$). The predicted values are very close to the measurements but the corresponding epistemic uncertainty is quite high. This can be explained by the fact that the training set contains less than $3\%$ of measurements $\geq 80mm$. \\
Example 2 is an example of propagation with decreasing values. The model underestimates crack lengths for the first few predicted values, then converges to the measured values at the end. However, falling values can be considered as inherent data noise or measurement errors, resulting in a high aleatoric uncertainty for this example. 

\section{Conclusion and future works}
\label{sec:conclusion}
Predicting the propagation of cracks in rails is a critical issue for optimizing the maintenance operations across the rail network. This task is intrinsically complex, and cannot be handled simply with physical simulations. In this paper, we proposed a deep learning approach based on real data collected on the rail. Obtained results show that the multi-horizons model outperforms conventional recurrent models such as GRU. The Bayesian multi-horizons model performs even better, and allows to quantify both aleatoric and epistemic uncertainties.
Several avenues of improvement can be investigated in future work, in particular the  calibration of  models to predict more accurate uncertainties, as proposed in \cite{kuleshov2018accurate}. We aim also at combining recurrent layers with attention layers that assign different weights to the hidden states based on their significance for forecasting the crack lengths. Finally, the hybridization of ML methods and physical simulations is also part of the work in progress. Indeed, information provided from physical simulation can contribute in enriching the variables of the learned model such as the wheel load of the vehicle rolling on the rail, and thus improving the prediction performance.  
\printbibliography

\end{document}